\documentclass[journal]{IEEEtran}

\usepackage{microtype}
\usepackage{graphicx}
\usepackage{amsmath}
\usepackage{booktabs} 

\usepackage{amssymb}
\usepackage{graphics, graphicx}
\usepackage{textcomp}
\usepackage{lmodern}
\usepackage{enumitem}

\usepackage{multirow}
\usepackage{adjustbox}

\usepackage{wrapfig,lipsum,booktabs}

\usepackage{algorithm}
\usepackage{algorithmic}

\usepackage{contour}

\usepackage{ulem}

\contourlength{0.8pt}

\newcommand{\myuline}[1]{%
  \uline{\phantom{#1}}%
  \llap{\contour{white}{#1}}%
}

\renewcommand{\bfseries}{\fontseries{b}\selectfont} 
\usepackage{etoolbox} 
\robustify\bfseries
\newrobustcmd{\B}{\bfseries}
\usepackage{makecell}

\usepackage{caption}
\usepackage{subcaption}
\usepackage{tikz}
\usepackage{url}
\usetikzlibrary{fit,positioning}

\newcolumntype{P}[1]{>{\centering\arraybackslash}p{#1}}
\newcolumntype{M}[1]{>{\centering\arraybackslash}m{#1}}

\newcommand{\quantities}[1]{%
  \begin{tabular}{@{}c@{}}\strut#1\strut\end{tabular}%
}

\newcommand{\ra}[1]{\renewcommand{\arraystretch}{#1}} 

\usepackage[utf8]{inputenc} 
\usepackage[T1]{fontenc}    
\usepackage{url}            
\usepackage{booktabs}       
\usepackage{amsfonts}       
\usepackage{nicefrac}       
\usepackage{microtype}      

\ifCLASSINFOpdf
\else
\fi
\begin{document}
%
\title{GRADE: Graph Dynamic Embedding}
%
%
%

\author{Simeon~Spasov,
        Alessandro~Di~Stefano,
        Pietro~Li\`o,
        Jian Tang 
\thanks {\textcopyright 20xx IEEE. Personal use of this material is permitted. Permission from IEEE must be obtained for all other uses, in any current or future media, including reprinting/republishing this material for advertising or promotional purposes, creating new collective works, for resale or redistribution to servers or lists, or reuse of any copyrighted component of this work in other works.}
\thanks{S.Spasov and P.Li\`o are with the Department
of Computer Science and Technology at the University of Cambridge, Cambridge, CB03DF, UK e-mail: (ses88@cam.ac.uk, pl219@cam.ac.uk).}
\thanks{A. Di Stefano is with Teeside University, Middlesbrough, TS1 3BX, UK.}
\thanks{Jian Tang is an assistant professor at Mila-Quebec AI Institute and HEC Montreal.}
\thanks{\textit{Corresponding author: Simeon Spasov}}}

%
%

\markboth{In submission to \textit{IEEE Trans. Neural Netw. Learn. Syst}}%
{Shell \MakeLowercase{\textit{et al.}}: Bare Demo of IEEEtran.cls for IEEE Journals}
%




\maketitle

\begin{abstract}
  Representation learning of static and more recently dynamically evolving graphs has gained noticeable attention. Existing approaches for modelling graph dynamics focus extensively on the evolution of individual nodes independently of the evolution of more global community structures. As a result, current methods do not provide useful tools to study and cannot explicitly capture temporal community dynamics. To address this challenge, we propose GRADE - a probabilistic model that learns to generate evolving node and community representations by imposing a random walk prior over their trajectories. Our model also learns node community membership which is updated between time steps via a transition matrix. At each time step link generation is performed by first assigning node membership from a distribution over the communities, and then sampling a neighbor from a distribution over the nodes for the assigned community. We parametrize the node and community distributions with neural networks and learn their parameters via variational inference. Experiments demonstrate GRADE outperforms baselines in dynamic link prediction, shows favourable performance on dynamic community detection, and identifies coherent and interpretable evolving communities.
\end{abstract}


%
\IEEEpeerreviewmaketitle

\section{Introduction}
Representation learning over graph-structured data has generated significant interest in the machine learning community owing to widespread application in a variety of interaction-based networks, such as social and communication networks, bio-informatics and relational knowledge graphs. Developing methods for unsupervised graph representation learning is challenging as it requires summarizing the graph structural information in low-dimensional embeddings. These representations can then be used for downstream tasks, such as node classification, link prediction and community detection. \\

The majority of unsupervised graph representation learning methods have focused solely on static non-evolving graphs, while many real-world networks exhibit complex temporal behaviour. To address the challenge of encoding temporal patterns of relational data, existing methods for dynamic graph embedding focus extensively on capturing node evolution \cite{goyal2018dyngem, goyal2020dyngraph2vec, sankar2020dysat, zhou2018dynamic}. Although these methods achieve compelling results against static baselines on dynamic tasks, they do not lend themselves to capturing the evolution of graph-level structures, such as clusters of nodes, or communities. On the other hand, the patterns of evolving node clusters are of great interest in social networks \cite{kossinets2006empirical, greene2010tracking, yang2011detecting}, as well as encountered in the temporal organization of large-scale brain networks \cite{vidaurre2017brain}, among others.\\

\begin{figure}[t]
\centering
\includegraphics[width=\linewidth]{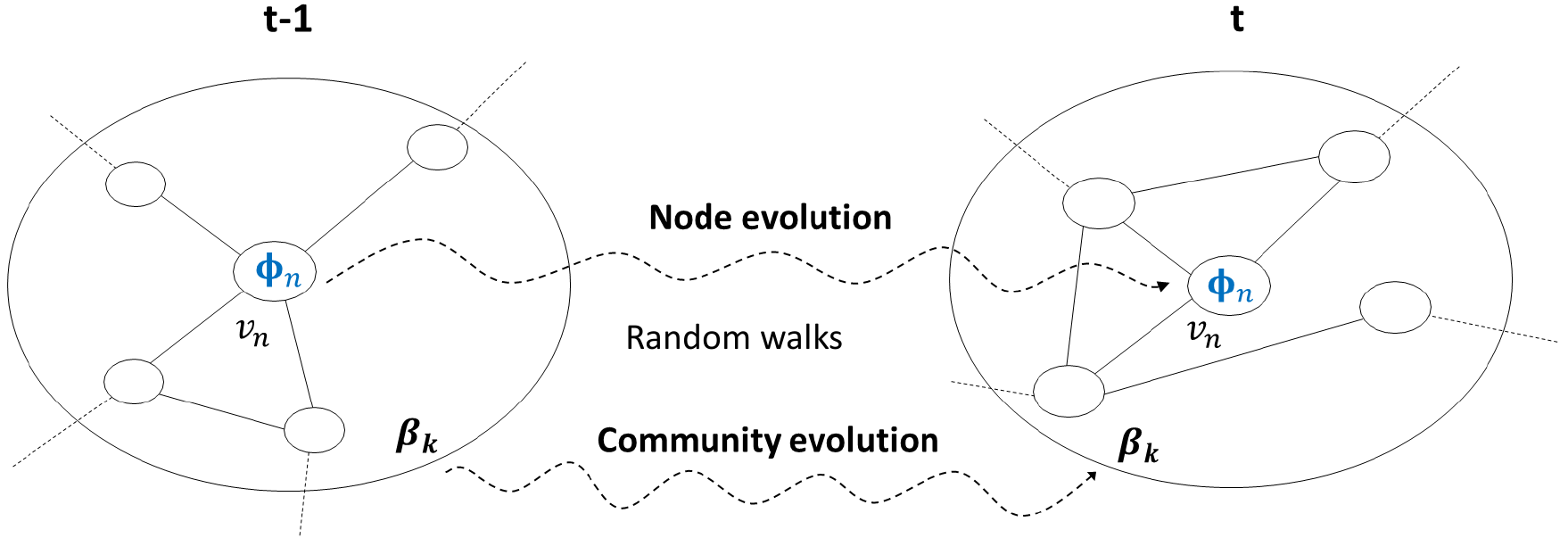}\quad
\caption{In our dynamic graph model, we assume that the semantic meaning of communities \textit{and} the social context of individual nodes evolve over time. This necessitates imposing temporal evolution on both the node and community embeddings, denoted by $\phi$ and $\beta$ respectively, between time steps, which we achieve via Gaussian random walks. }
\label{Node_and_community_evolution}
\end{figure}

To address this challenge, we propose GRADE (GRAph Dynamic Embedding) -  a probabilistic generative model for jointly learning evolving node and community representations. The benefit of modelling the interaction between nodes and communities for graph representation learning in the static setting was studied 
by vGraph \cite{vGraph}. Further, \cite{battiston2020networks} produces evidence that taking into consideration higher-order graph structures, such as communities, enhances our capability to model emergent dynamical behaviour. Consequently, in this work, we extend the idea of modelling node-community interactions to the dynamic case. We represent a dynamic network as a sequence of graph snapshots over a series of discrete and equally-spaced time intervals. Similarly to vGraph, at each time step, we model the edge generation process between node neighbours via multinomial community and node distributions. First, we sample a community assignment $z$ for each node $v$ from a distribution over the communities, i.e. $z\sim p(z|v)$.
Then, we sample a neighbour $c$ from the distribution over the nodes of the assigned community, that is $c \sim p(c|z)$. Both the community and node distributions are parametrized by neural network transformations of the node and community embeddings. In our work, we assume that the semantic meaning of communities \textit{and} the proportions over the communities for each node evolve simultaneously over time.
Following an approach introduced in dynamic topic modelling \cite{dieng2019dynamic}, we encode temporal evolution in our method by assuming a random walk prior over the representations between time steps (see Fig. \ref{Node_and_community_evolution}). Furthermore, we draw inspiration from social networks where a user's preferences can shift from one community to another. 
We explicitly model the dynamism in community membership by introducing a node-specific and time-varying transition matrix to update the community mixture coefficients over time. We design an effective algorithm for inference via backpropagation. We learn the parameters of our model by means of variational inference to maximize the lower bound of the observed data. More specifically, we resort to amortized inference \cite{gershman2014amortized} to learn neural network mappings from node and community representations to the respective conditional distributions, as well as structured variational inference \cite{hoffman2015structured, saul1996exploiting} to retain the dependence of the embeddings on their historical states. Our proposed method is aimed at non-attributed dynamic graphs. It is worth noting that although GRADE is a transductive approach, changes of vertex sets between snapshots at different time steps do not pose a problem, if the complete vertex set is known a priori. Our main contributions can be summarized as follows:
\vspace{1em}
\begin{itemize}
    \itemsep1em
    \item [1)] We introduce \textit{node and community dynamics} by evolving their respective representations as \textit{Gaussian random walks}. We also explicitly update the community mixture coefficients via a \textit{transition matrix}.
    \item [2)] We use learnable transformations of the node and community embeddings, that is neural networks, to produce the parameters of the node and community distributions in the implementation of our dynamic graph generative model.
    \item [3)] We propose  an efficient variational inference procedure to learn the parameters of the model. The use of neural networks to implement the generative model (point 2.) \textit{facilitates the use of amortized inference}. We only need to augment the input to the neural networks in order to generate the parameters of the posterior node and community distributions. Further, owing to our choice of the posterior family, we can easily retain the dependence of the node and community embeddings on their historical states (structured variational inference), which \textit{improves the fidelity of our approximation to the true posterior}.
    \item [4)] We show our method \textit{outperforms} both static and dynamic baselines on the tasks dynamic link prediction and dynamic community detection on real-world dynamic graphs extracted from the DBLP bibliography, Wikipedia and Reddit. More specifically, as a probabilistic model, GRADE allows us to \textit{explicitly infer community assignments}, whereas competitive models for dynamic graph representation learning do not possess this capacity. As a result GRADE shows very strong results on the task of dynamic community detection.
\end{itemize}
\vspace{1em}
Finally, GRADE is also practical to train. Our method takes up to $\sim12$ hours on a single 12GB NVIDIA TitanX GPU for a dynamic graph with over $35,000$ nodes and $180,000$ edges.

\section{Related Work}

\begin{figure}[t]
\begin{center}
\centerline{\includegraphics[width=1.\columnwidth]{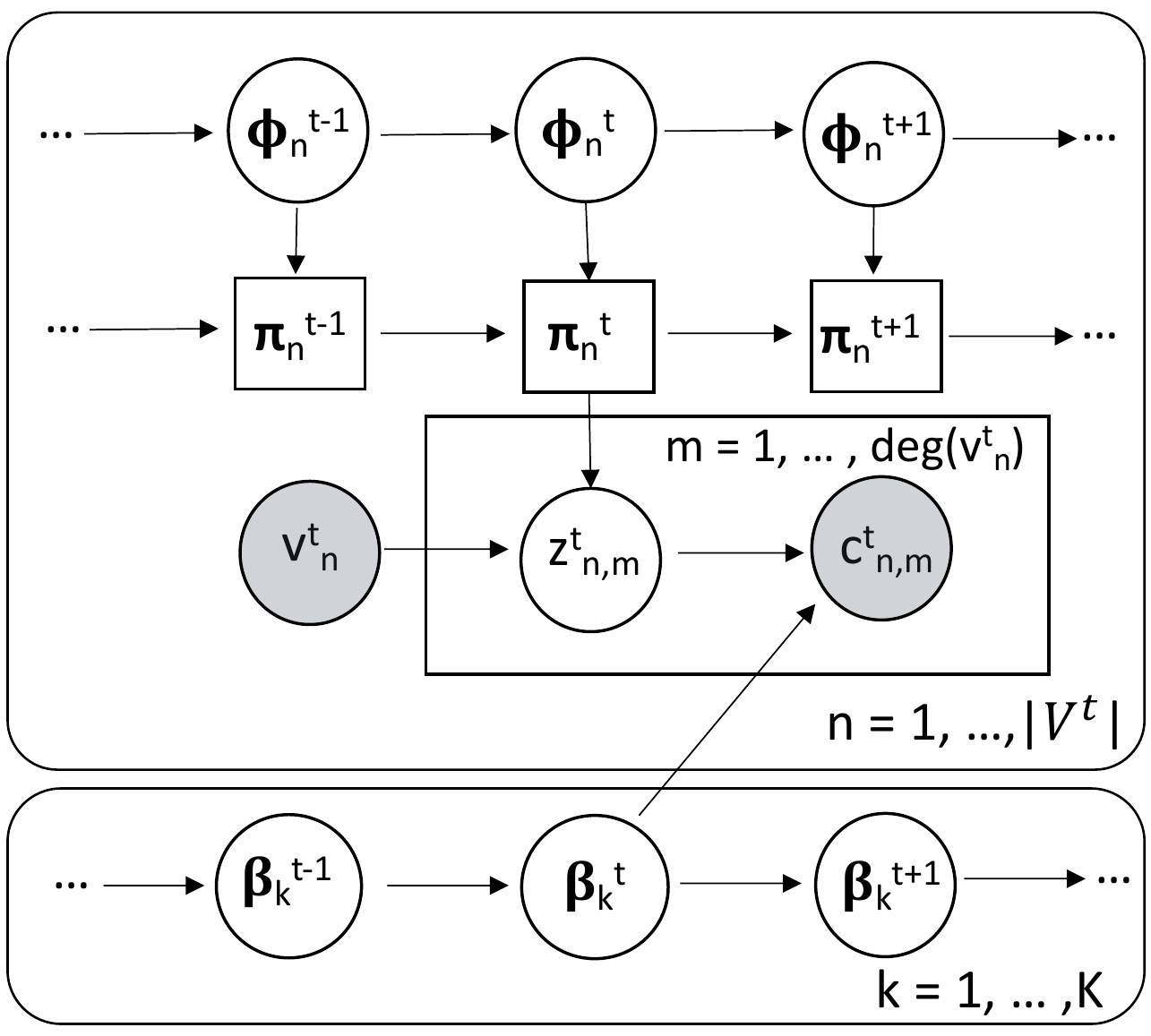}}
\caption{Plate notation for the edge generative process of GRADE. The node and community representations, denoted by $\phi$ and $\beta$ respectively, and consequently the parametrization of the node and community distributions, evolve over time. The parameters of the community distribution, $\pi$, are explicitly updated by a deterministic transformation  of the node embeddings (denoted by rectangules). The observed data is the edges ($v_n$, $c_{n,m}$) in the dynamic graph.
}
\label{fig2}
\end{center}
\end{figure}

Methods for unsupervised learning on evolving graphs are often dynamic extensions of ideas applied in the static case. (1) \myuline{Graph factorization} approaches such as DANE \cite{li2017attributed} rely on spectral embedding, similarly to static methods like \cite{ahmed2013distributed, cao2015grarep, belkin2003laplacian}. DANE assumes smooth temporal evolution and models it using matrix perturbation theory. (2) In \myuline{skip-gram models} node representations are learnt by random walk objectives \cite{grover2016node2vec, perozzi2014deepwalk, tang2015line}. In the dynamic case, CTDNE \cite{nguyen2018continuous} and NetWalk \cite{yu2018netwalk} augment the random walk with temporal constraints based on time-stamped edges.
Further, (3) \myuline{temporal point processes} have also been used in combination with neural network parametrization by KnowEvolve \cite{trivedi2017know} and DyREP \cite{trivedi2018dyrep} to model continuous-time node interactions in multi-relational and simple dynamic graphs, respectively. More recently, (4) \myuline{graph convolutional neural networks} (GNNs) have become a widely used tool for graph representation learning \cite{kipf2016variational, velivckovic2018deep, bruna2014}. A popular approach to include temporal dependency in GNNs is to introduce a recurrent mechanism. For example, \cite{seo2018structured} proposes two ways to achieve this goal. One way is to obtain node embeddings via a GNN, which are then fed to an LSTM to learn dynamism. 
The second approach modifies the LSTM layers to incorporate graph-structured data. A different approach altogether is to evolve the graph convolutional parameters with a recurrent neural network (RNN), such as in EvolveGCN \cite{pareja2020evolvegcn}, as opposed to the node embeddings, hence addressing issues stemming from rapidly changing node sets between time steps. Alternatively, STGCN \cite{yu2017spatio} avoids using RNNs completely by introducing an efficient ST-Conv layer for faster training with few model parameters.\\

The most related body of work to this paper is a category of transductive unsupervised methods applied on temporally discrete and non-attributed dynamic graphs. One such approach is DynGEM \cite{goyal2018dyngem} which employs deep autoencoders to learn node embeddings but resorts to no recurrent structures for temporal dependency. Instead, time dynamics are injected by re-initializing the parameters of the autoencoder at each time step $t$ with the parameters learnt at $t-1$. Unlike our proposed method GRADE, DynGEM can only handle growing dynamic networks. Another method is dyngraph2vec \cite{goyal2020dyngraph2vec} which is trained to predict future links based on current node embeddings using an LSTM mechanism. DynamicTriad \cite{zhou2018dynamic} models the process of temporal link formation between vertices with common neighbours, that is triadic closure, and enforces latent space similarity between future connected nodes. Finally, DySAT \cite{sankar2020dysat} draws inspiration from the success of attention mechanisms and applies them structurally over local neighbourhoods and temporally over historical representations. It is important to note that DySAT has been demonstrated to consistently outperform other dynamic graph representation learning methods, more specifically DynGEM, Dynamic Triad and dyngraph2vec, and hence serves as our strongest dynamic baseline.\\

The main advantage of GRADE over these competitive methods is that GRADE learns evolving community \textit{as well as} node representations, thus enabling us to explicitly capture community dynamics. This also allows for an effective mechanism for community detection, which we demonstrate in our experiments. Further, our approach can be used to infer the embeddings for future time steps. In comparison, these dynamic methods use representations learnt at the last training step for dynamic prediction.\\

Finally, GRADE is also related to dynamic topic modelling \cite{dieng2019dynamic, blei2006dynamic} as both can also be viewed as state-space models. The difference is that in GRADE we are dealing with multinomial distributions over the nodes and communities instead of topics and words. Moreover, some works like \cite{bamler2017dynamic, rudolph2018dynamic} have focused on the shift of word meaning over time, and others such as \cite{dieng2019dynamic} model evolution of topic proportions in documents. In contrast, GRADE assumes both nodes and communities undergo temporal semantic shift. 

\section{Problem Definition and Preliminaries}

\begin{table}[t] \centering
\caption{Notation commonly used in this paper.}
\ra{1.9}
\vspace{5pt}
    \label{wrap-tab:notation}
  \begin{tabular}{lc} \toprule 
    \bf Symbol     & \bf Definition   \\ 
    \midrule
    $\Gamma = G^{1}, \ldots, G^{T}$  & sequence of dynamic graph snapshots \\
    $v_{n} \in \mathcal{V}=\{v_1,\ldots,v_N\}$ & complete vertex set of $\Gamma$\\
    $k={1,\ldots, K}$  & community indices \\
    $\beta, \phi$  & community and node embeddings \\
    $\gamma, \sigma$ & temporal smoothness hyperparameters\\
    $z$ & community assignment latent variable\\
    $A_{n}^{t}$ & transition matrix for node $v_n$ at time $t$ \\
    edge $(v,c)$ & (source node, target node) \\
    $n=1, \ldots, \vert V^{t} \vert$ & vertex set indices of $G^{t}$ \\
    $m = 1, \ldots, deg(v_{n})$ & neighbour index of $v_{n}$ \\
    $\pi, \theta$ & \quantities{parameters of node and \\ community distributions} \\
    \bottomrule
  \end{tabular}
    \end{table}

We consider a dataset comprising a sequence of non-attributed (i.e. without node features) graph snapshots $\Gamma = G^{1}, \ldots, G^{T}$ over a series of discrete and equally-spaced time intervals $t \in \{1,\ldots,T\}$, such that $t$ is an integer time index.  We assume all the edges in snapshot $G^t$ occur at time $t$ and the complete set of vertices $\mathcal{V}=\{v_1,\cdots,v_N\}$ in the dynamic graph $\Gamma$ is known a priori. Our method supports the addition or removal of nodes as well as edges between time steps. We also assume there are $K$ communities (clusters of nodes) in the dynamic network. Our method aims to learn time-evolving vector representations $\phi_{n}^t \in R\textsuperscript{\textit{L}}$ for all nodes $v_n \in \mathcal{V}$ and communities $\beta_{k}^t \in R\textsuperscript{\textit{L}}$, $k \in \{1,\ldots, K$\} , for each time step $t$. Further, a useful model for dynamic graph embedding should not only capture the patterns of temporal evolution in the node and community representations but also be able to predict their future trajectories.

\section{Methodology}
\subsection{GRADE: Generative Model Description}

\begin{figure*}[t]
\centering
\includegraphics[width=\linewidth]{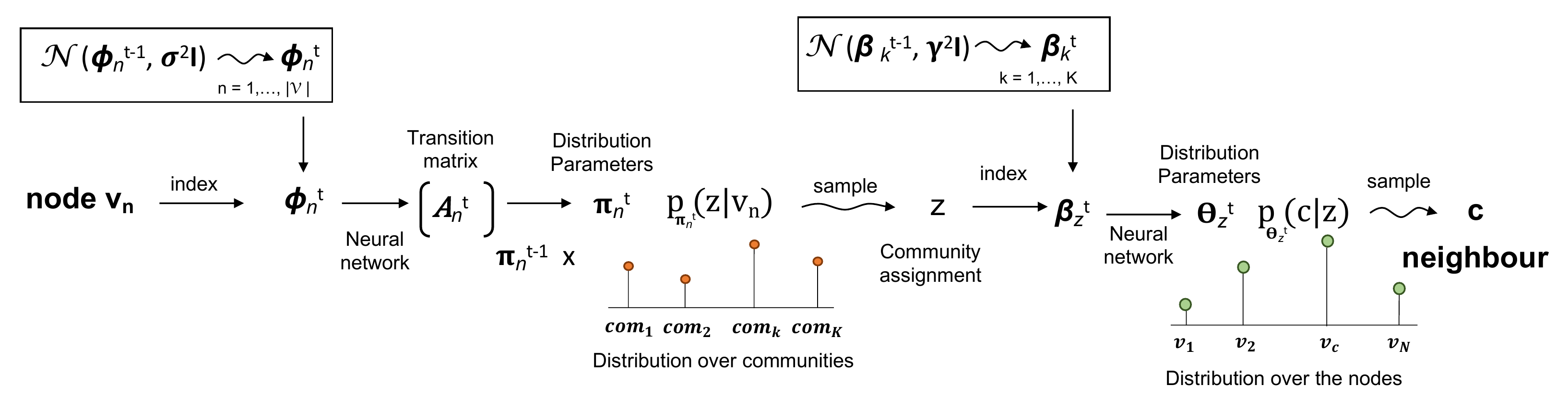}\quad
\caption{In this figure we present GRADE's edge generative process for a source and target node pair at time step $t$. Given a source node $v_n$, we sample its embedding, $\phi_n$, from its prior distribution. Then, we transform this embedding into a community transition matrix, which we use to update the parameters of the multinomial distribution over the communities for node $v_n$. We sample a community assignment $z$ from the community distribution of $v_n$. Then, we use the corresponding community embedding $\beta_z^t$, and use it to parameterize a multinomial distribution over the nodes for community $z$ after a neural network transformation. Finally, we sample a node neighbour $c$. }
\label{Generative_flowchart}
\end{figure*}

\vspace{0.5em}
GRADE is a probabilistic method for modelling the edge generation process in dynamic graphs. Similarly to vGraph~\cite{vGraph}, we represent each node $v$ in the active vertex set of $G^t$ as a mixture of communities, and each community as a multinomial distribution over the nodes. The linked neighbour generation at each time step $t$ is as follows: first, we sample a community assignment $z$ from a conditional prior distribution over the communities $z \sim p(z \vert v)$. Then, a neighbour is drawn from the node generative distribution $c \sim p(c \vert z)$ based on the social context defined by the assigned community. The generative process for graph snapshot $G^t$ can be formulated as:
\vspace{6pt}
\begin{equation}
p(c \vert v) = \sum_{z}p_{\theta^{t}}(c\vert z)p_{\pi^{t}}(z \vert v) 
\label{eq: generative model}
\end{equation}

where $\theta^t$ and $\pi^t$ parametrize the multinomial generative and prior distributions at time step $t$ respectively. In our dynamic graph model, we suppose that the semantic meaning of communities \textit{as well as} community proportions for nodes change over time. This necessitates capturing the temporal evolution of the underlying node and community distributions by an evolving set of distribution parameters.
GRADE achieves this by making the distribution parameters implicitly dependent on the evolving node and community embeddings, $\phi$ and $\beta$ respectively.\\

More specifically, we treat the community and node representations as random variables and impose a simple state-space model that evolves smoothly with Gaussian noise between time steps as follows:
\vspace{6pt}
\begin{equation}
p(\beta_{k}^t \vert \beta_{k}^{t-1}) = \mathcal{N}(\beta_{k}^{t-1}, \gamma^2I)
\label{eq: beta evolve}
\vspace{6pt}
\end{equation}
\begin{equation}
p(\phi_{n}^t \vert \phi_{n}^{t-1}) = \mathcal{N}(\phi_{n}^{t-1}, \sigma^2I)
\label{eq: phi evolve}
\vspace{6pt}
\end{equation}

Note that we evolve the embeddings of the complete vertex set $\mathcal{V}$ at each time step, although our model allows for a subset of the nodes to be present at each time step. Assuming a Gaussian state-space temporal model for the evolution of continuous variables, such as the node and community representations in GRADE, is common in Hidden Markov Models (HMMs) \cite{ghahramani2001introduction}. In general, temporal dynamics in HMMs can be decomposed into a deterministic and a stochastic component. Using $\phi$ as an example: $\phi_n^t = f^t(\phi_n^{t-1}) + \epsilon$, where $t$ is a discrete time index, $f^t$ is a deterministic function and $\epsilon$ is a zero-mean random noise vector. In Eq. \ref{eq: beta evolve} and Eq. \ref{eq: phi evolve}, for the deterministic component, $f^t$, \textit{we choose the identity matrix}. We could make more complex design choices, for example making $f^t$ a transformation learnt from the data. Using the identity matrix showed strong results so we did not explore other approaches, which would also introduce more complexity. We sample the $\epsilon$ noise vector from  $\mathcal{N}(0, \gamma^2I)$ and $\mathcal{N}(0, \sigma^2I)$, respectively, where $\sigma$ and $\gamma$ are temporal smoothness hyperparameters. By tuning the smoothness hyperparameters, we have direct control over the assumed rate of temporal dynamics, hence also implicitly control the temporal smoothness of the node and community distribution parameters. \\

At each time step $t$, the parametrization of the community distributions over the nodes is achieved by first transforming the community representations through a neural network, and then mapping the output through a softmax layer: $\theta_k^t = $softmax$(\zeta(\beta_k^t))$. To evolve the parameters of the node distributions, we observe that users' interests in a social network change over time. As a result, users may shift from one community to another. This is characterized by user-specific behaviour within the broader context of community evolution. For these reasons, we explicitly model community transition with a transition matrix.\\

\begin{figure*}[t]
\centering
\includegraphics[width=\linewidth]{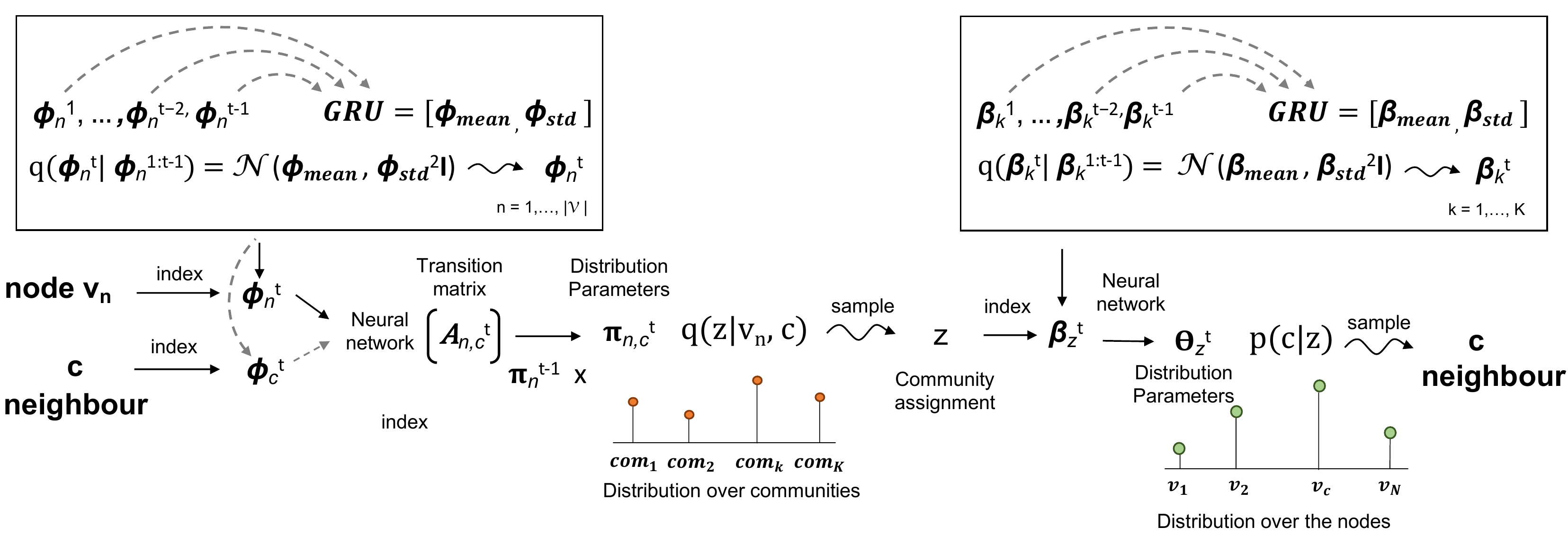}\quad
\caption{We present the inference algorithm we propose to learn the parameters of GRADE. The grey dashed lines denote variable dependencies that are necessary for the variational approximations of the intractable posteriors. The temporal dependency of the node and community embeddings is captured with Gated Recurrent Units (GRUs), which use all previous historical states to produce the mean and covariance vectors of their respective Gaussian variational approximations. In order to retain the dependency of the approximation over the communities, $q(z|v_n, c)$, on both the source and target nodes, we use both source and target node embeddings as inputs to the neural network which produces the community transition matrix.}
\label{Inference_flowchart}
\end{figure*}

More specifically, for each node $v_n$, we update the community mixture weights $\pi_n^{t}$ by means of a $K\times K$ node-specific and time-varying transition matrix $A_n^{t}$, produced as a function, $\psi$, of the node embeddings:
\begin{equation}
\pi_n^{t} = \pi_n^{t-1} \cdot A_n^{t} 
\label{eq: pi evolve}
\end{equation}
Note that $\pi$ is a row vector. Another advantage for introducing an explicit transition matrix is the option to impose, for instance, a diagonal dominance prior if we believe community transitions happen infrequently. Although this investigation is left as future work, it justifies the choice of introducing a transition matrix to update the community mixture coefficients.\\

In summary, GRADE's edge generative process for each graph snapshot $G^t$ in $G^1, \ldots, G^T$ is as follows: 
\begin{enumerate}
    \itemsep0.5em
	\item Draw community embeddings $\beta_{k}^t$ for $k = 1,\cdots, K$: 
	   \hspace*{0.4em}$\beta_{k}^t \sim \mathcal{N}(\beta_{k}^{t-1}, \gamma^2I)$
	\item Draw node embeddings $\phi_{n}^t$ for all nodes $v_n \in \mathcal{V}$: \hspace*{0.4em}$\phi_{n}^t \sim \mathcal{N}(\phi_{n}^{t-1}, \sigma^2I)$
	\item Transition matrix $A_n^{t}$ is a non-linear transformation, $\psi$, of the node embeddings:\\ \hspace*{0.3em}$A_n^{t}= $ row-softmax $(\psi(\phi_n^{t}))$
	\item Update community mixture coefficients for node $v_n$: \hspace*{0.2em}$\pi_n^{t}=\pi_n^{t-1} \cdot A_n^{t}$ 
	\item For each edge ($v_n^t$, $c_{n,m}^t$) in $G\textsuperscript{t}$: 
	\begin{enumerate}
    	\item Draw community assignment $z$ from multinomial prior over the communities:\\ \hspace*{0.3em}$z_{n,m}^t \sim p_{\pi_n^t}(z \vert v_n^t)$ 
        \item Parameters of distribution over the nodes is a function of $\beta_z^{t}$: \hspace*{0.3em}$\theta_z^{t} =$ softmax$(\zeta(\beta_z^{t}))$
        \item Draw linked neighbour $c$ from node generative distribution for sampled community $z$: \\
        \hspace*{0.3em}$c_{n,m}^t \sim p_{\theta_z^t}(c \vert z_{n,m}^t)$
\end{enumerate}
\end{enumerate}
\vspace{1em}
The graphical model of the proposed generative process is shown in Fig. \ref{fig2} and common notation used in the paper in Table \ref{wrap-tab:notation}. One \textit{key difference} between vGraph and GRADE is that vGraph uses the multiplication of the node and community embeddings to produce the distribution parameters, that is softmax($\phi_n^{T}\beta_k$). In GRADE we cannot employ this approach because the edge generative mechanism has been changed to accommodate for temporal dynamics. Firstly, for each node, we rely on a transition matrix to update the community mixture coefficients. The transition matrix is a (neural network) function \textit{solely} of the node embeddings since it is node-specific. Second, in order to produce the parameters of the node distribution (for a given community), we choose to use a neural network transformation of the community embeddings. 
\subsection{Inference Algorithm}
Consider we are given a dataset $\mathcal{D}$ comprising a set of node links ($v_n^t$, $c_{n,m}^t$) for a sequence of graph snapshots $G\textsuperscript{1}, \ldots, G\textsuperscript{T}$. In our dynamic graph model, the latent variables are the hidden community assignments $z_{n,m}^t$, and the evolving node and community representations $\phi_n^t$ and $\beta_k^t$. The logarithm of the marginal probability of the observations is given by the sum of the log probability of each observed temporal edge $\log p(c_{n,m}^t \vert v_{n}^t)$ 
of all nodes in the dynamic graph $\Gamma$:
\begin{equation}
\textrm{log}p(\mathcal{D}) = \sum_{t=1}^{T} \sum_{n=1}^{\vert V^t \vert} \sum_{m=1}^{deg(v_n^t)} \textrm{log}p_{\theta^t,\pi^t}(c_{n,m}^t\vert v_n^t)
\label{eq: Eq. 7}
\end{equation} 
Exact inference of the posterior $p(\phi, \beta, z \vert \mathcal{D})$ is intractable. Instead, we resort to variational methods as a means of approximation. Variational inference provides flexibility when choosing an appropriate family of distributions $q(\phi, \beta, z \vert \mathcal{D})$ as an approximation to the true posterior. Choosing such a family of variational distributions which is sufficiently flexible and adheres to the conditional dependencies of the hidden variables while allowing for an efficient optimization procedure is one of the main challenges in variational inference. The variational approximation we choose for our model takes the form:
\begin{equation}
\begin{split}
q(\phi, \beta,z)&=\prod_{t=1}^{T} \prod_{n=1}^{\vert \mathcal{V}  \vert}q(\phi_n^t\vert\phi_n^{1:t-1}) \\
&\times \prod_{t=1}^{T}\prod_{k=1}^{K}q(\beta_k^t\vert\beta_k^{1:t-1}) \\
&\times \prod_{t=1}^{T}\prod_{n=1}^{\vert V^t\vert} \prod_{m=1}^{deg(v_n^t)}q(z \vert v_n^t, c_{n,m}^t)
\label{eq: var_approx}
\end{split}
\end{equation}

For ease of implementation, in the factorization of the approximated posterior, we have kept the individual hidden variables $\phi, \beta, z$ independent, although we retain the temporal dependency of the node and and community embeddings on their historical states. \\ 

The goal of variational inference is to optimize the parameters of the approximated posterior by minimizing the Kullback-Leibler (KL) divergence between the true posterior and its approximation. This procedure is known as minimizing the evidence lower bound (ELBO)
\cite{kingma2013auto}:

\vspace{2pt}
\begin{equation}
\mathcal{L}=E_q[\textrm{log} p(\mathcal{D},\phi,\beta,z)-\textrm{log}q(\phi, \beta,z)]
\label{eq: elbo}
\end{equation}
\vspace{1pt}

The ELBO for GRADE can be formulated as a neighbour reconstruction loss, and a sum of KL regularization terms between the priors and posteriors of each of the latent variables. 
At each time step $t$, the ELBO from Eq.\ref{eq: elbo} can be expressed as: 

\begin{equation}
\begin{split}
& \mathcal{L}^t = E_{z^t\sim q(z|v^t, c^t)}[\textrm{log}p_{\theta^t}(c|z^t)] - \\[2.5ex]
&\hspace*{35pt} KL[q(z|v^t, c^t)||p_{\pi^t}(z|v^t)] - \\[2.5ex]
&\hspace*{35pt} \sum_k KL[q(\beta_k^t|\beta_k^{1:t-1})||p(\beta_k^t|\beta_k^{t-1})] - \\
&\hspace*{35pt}\frac{1}{\mathcal{|V|}} \sum_n^{|\mathcal{V}|} KL[q(\phi_n^t|\phi_n^{1:t-1})||p(\phi_n^t|\phi_n^{t-1})]
\end{split}
\label{eq: elbo_detailed}
\end{equation}
\vspace{1pt}

We omit denoting the community and node indices for notational clarity in Eq. \ref{eq: elbo_detailed}. Note that we scale down the contribution of the KL divergence loss between the prior and posterior for the node representations by multiplying by $\frac{1}{\mathcal{|V|}}$, that is the size of the vertex set. We find that averaging over all nodes prevents from overpowering the other loss terms. \\

\setlength{\textfloatsep}{12,5pt}
\begin{algorithm}[t]
   \caption{GRADE Inference Algorithm}
   \label{alg:1}
\begin{algorithmic}
   \STATE {\bfseries Input:} Edges $(v_{n}^t,c_{n,m}^t)$  in dynamic graph $\Gamma=G^1,\ldots, G^T$.
 \STATE Initialize all generative and variational parameters
 \STATE Initialize $\beta^0$, $\phi^0$ as learnable parameters. $\pi^{0}$ is uniform over $K$.
   \FOR{ iterations 1, 2, 3, \ldots}
  \FOR{ graph $G^t$ in $G^1, \ldots, G^T$}
  \STATE $\diamond$ \textbf{\small{Sample community representations from posteriors}}
   \FOR{ k in $1, \ldots, K$}
   \STATE $\beta_k^{t} \sim q(\beta_k^{t} \vert \beta_k^{1:t-1})
   $
   \ENDFOR
   \STATE $\diamond$ \textbf{\small{Sample node representation for complete vertex set from posteriors}}
   \FOR{ n in 1, \ldots, |$\mathcal{V}$|}
   \STATE $\phi_n^{t} \sim 
   q(\phi_n^t \vert \phi_n^{1:t-1})$
   \ENDFOR
   \STATE $\diamond$ \textbf{\small{Sample community assignments for edges}}
   \FOR{edges ($v_{n}^t,c_{n,m}^t$) in $G^{t}$ }
    \STATE $z_{n,m}^t \sim q(z \vert v_{n}^t,c_{n,m}^t)$
    \STATE $c_{n,m}^t\sim p_{\theta_z^t}(c \vert z_{n,m}^t)$
   \ENDFOR
   \ENDFOR
   \\Estimate ELBO (eq.~\eqref{eq: elbo_detailed}) and update the generative and variational parameters via backpropagation.
   \ENDFOR
\end{algorithmic}
\end{algorithm}

The variational distributions over the node and community representations depend on all of their respective historical states. I capture this temporal dependency with Gated Recurrent Units (GRU). Note that any other recurrent mechanism (e.g. Long short-term memory networks) can be used for this purpose, as well as temporal attention, such as in \cite{sankar2020dysat}. I model the outputs of both $q(\beta_k^t\vert\beta_k^{1:t-1})$ and $q(\phi_n^t\vert\phi_n^{1:t-1})$ as Gaussian distributions, similarly to their priors. In the posteriors however, the means and diagonal covariance vectors are given by the outputs of their respective GRU units. This approach in which some of the dependency of the hidden variables is retained, in this case over their historical states, is known as structured variational inference \cite{hoffman2015structured,saul1996exploiting}. There are several advantages to this approach: 1) it allows to easily infer the posterior distributions of the node and community representations at future time steps by running the GRUs after training; 2) I explicitly retain the historical dependence of the node and community embeddings, which increases the fidelity of the approximation to the true posterior; 3) the number of model parameters introduced by the GRU modules are negligible. Also, the number of model parameters does not scale with time, but does scale with the number of nodes and communities. \\

The difference between the prior over the community assignments $p_{\pi}(z|v)$ and the approximated posterior $q(z|v,c)$ is in the dependence on the neighbour $c$. In principle, this dependency can be easily integrated in the parametrization via amortized inference \cite{gershman2014amortized}. More specifically, I use both embeddings of the source, $v$, and target nodes, $c$, as inputs to the transformation $\psi$ generating the community transition matrix.  This introduces no new model parameters as the same $\psi$ function can simply be reused. Also, the structure of the variational distribution over the assignments $z$ enables an efficient procedure for inferring edge labels, as well as community memebership approximation as follows:

\begin{equation}
p(z \vert v) \approx \dfrac{1}{|N(v)|}\sum_{c\in N(v)}q(z|v, c)
\label{eq: Eq. comm_membership_eq}
\vspace{2pt}
\end{equation}

where $N(v)$ is the set of neighbors of node $v$. The procedure is also applicable on future test graphs. Optimizing the  lower bound (eq.~\eqref{eq: elbo}) w.r.t. all parameters is performed based on stochastic optimization using the reparametrization trick \cite{kingma2014semi} and Gumbel-Softmax reparametrization \cite{jang2016categorical,maddison2016concrete} to obtain gradients. Refer to Algorithm~\ref{alg:1} and Fig. \ref{Inference_flowchart} for a summary of the inference procedure. \\

To implement GRADE, I use GRUs (one for the node and community evolution respectively) with a single layer. For the $\psi$ and $\zeta$ functions, which are used to generate the parameters of the distributions over the communities and nodes respectively, I use simple linear layers. These are all the components required.

\begin{table}[t] \centering
\caption{Datasets statistics. We use the proportion of time steps a node is present in to measure average node activity. The rate of context dynamics is captured by the Jaccard coefficient between the sets of immediate neighbours in consecutive time steps across all active nodes (lower coefficient suggests high rate of context dynamics).}
  \ra{1.5}
    \vspace{10pt}
    \begin{adjustbox}{width=\linewidth}
  \begin{tabular}{lccc} \toprule
        & \bf DBLP     & \bf Reddit  & \bf Wikipedia\\ \midrule
     \# Nodes & 10,000 & 35,776 & 5,000 \\ 
    \# Links & 374,911 & 180,662 & 482,069 \\
    Node Activity & 0.47 & 0.25 & 0.76\\
    Context Dynamics & 0.30 & 0.24 & 0.13 \\
    Label Rate & 0.083 & - & - \\
    Train/Val/Test & 13/3/3 & 6/2/2 & 7/3/3 \\
    snapshots &  &  &  \\
    \bottomrule
  \end{tabular}
  \end{adjustbox}
  \label{dataset_statistics}
\end{table}

\section{Experiments}
\label{sec:experiment}
We evaluate our proposed model on the tasks of dynamic link prediction and dynamic community detection against state-of-the-art baselines. Furthermore, we propose a quantitative metric to assess the quality of the learnt evolving communities and provide visualizations for a qualitative assessment. 
\subsection{Data sets}
\label{ssec:dataset}
We use three discrete-time dynamic networks based on the DBLP, Wikipedia and Reddit datasets to evaluate our method. A summary of all datasets is provided in Table~\ref{dataset_statistics}.\\

\textbf{DBLP} is a computer science bibliography which provides publication and author information from major journals and conference proceedings. We pre-process the DBLP dataset to identify the top 10,000 most prolific authors in terms of publication count in the years 2000-2018 inclusive. For each year, we construct a graph snapshot in which we connect author nodes if they have co-authored a publication in the same year. We stratify the DBLP dataset in 8 sub-fields, that is communities, based on publication venue. The communities comprise Artificial Intelligence, Computational Linguistics, Programming Languages, Data Mining, Databases, Systems, Hardware, Theory. We produce yearly labels for authors if over half of their annual publications fall within the same research category.\\

\textbf{Reddit} is a timestamped hyperlink network between subreddits \cite{kumar2018community}. An edge in the dataset represents a subreddit comment posting a hyperlink to another subreddit. Each edge has a timestamp associated with it. The complete dataset spans 2.5 years from January 2014 to April 2017 (overall 40 months). We divide the edges in 4-month blocks to create 10 graph snapshots spanning 40 months. \\

\textbf{Wikipedia} is a temporal network representing users editing each other's Talk pages \cite{kumar2018community, paranjape2017motifs}, and spans data between years 2002 and 2007 inclusive. An edge ($v, c, t$) means that user $v$ edited user $c$'s page at time $t$. During pre-processing, we first identify the 5,000 most active editors in the complete dataset. Then, we divide the edges for years 2006 and 2007 in 8-week periods (13 periods in total). The edges for each of the periods comprise a single graph snapshot.

\subsection{Baseline methods}
\label{ssec:baseline}
We compare GRADE against five baselines comprising three static and two dynamic methods. The static methods are: \textbf{DeepWalk} \cite{perozzi2014deepwalk}, \textbf{node2vec} \cite{grover2016node2vec} and \textbf{vGraph} \cite{vGraph}. The dynamic graph methods consist of: \textbf{DynamicTriad} \cite{zhou2018dynamic} and \textbf{DySAT} \cite{sankar2020dysat}.

\begin{table}[t] \centering
    \caption{Mean average rank (MAR) results on dynamic link prediction. Lower values are better. \textbf{Best} and \underline{second-best} results are marked in bold and underlined respectively. GRADE outperformes all baselines methods significantly on all datasets.}
     \vspace{21.5pt}
    \ra{1.5}
    \begin{adjustbox}{width=\linewidth}
        \begin{tabular}{lccc} \toprule
           & \bf DBLP & \bf Reddit  & \bf Wikipedia \\ \midrule
        DeepWalk  & $1,757 \pm 1$ & $4,322 \pm 2$ & $1,223  \pm 1.2$ \\
        Node2Vec  & $2,418 \pm  9$ & $6,863 \pm 28$ & \underline{1,131 $\pm$ 2} \\
        DySAT & $1,505 \pm  0.1$ & $4,040  \pm 5$ & $1,546  \pm 1$ \\
        DynTriad & $1,905 \pm  12$ & $1,575  \pm 36$ & $1,390  \pm 13$ \\
        \bf GRADE & \bfseries{605 $\pm$  39} & \bfseries{601  $\pm$ 8} & \bfseries{343  $\pm$ 6} \\
        vGraph & \underline{1,436 $\pm$  33} & \underline{1,223  $\pm$ 33} & $1,439  \pm 3$ \\
        \bottomrule
        \end{tabular}
        \end{adjustbox}
        \label{mars}
\end{table}


\begin{table*}[!ht]
  \centering
  \caption{Dynamic community detection performance. \textbf{Best} and \underline{second-best} results are marked in bold and underlined respectively. Values within a standard deviation on the same task are both marked.}
  \vspace{15pt}
  \ra{1.5}
  \begin{adjustbox}{width=\linewidth}
  \begin{tabular}{l c c c c c c c c}
    \toprule
     Measure & Dataset & DeepWalk & Node2Vec & DySAT &
     DynTriad &
     GRADE &
     vGraph  \\
    \hline
    \multirow{4}{*}{Modularity} & DBLP & $0.295 \pm 0.002$ & $0.314 \pm 0.002$ & $0.306 \pm 0.0$ & $0.188 \pm 0.001$ & \bfseries{0.383 $\pm$ 0.002} & \underline{$0.374 \pm 0.001$} \\
    & Wikipedia & $0.115 \pm 0.001$ & $0.113 \pm 0.001$ & $0.071 \pm 0.0$ & $0.092 \pm 0.001$ & \bfseries{0.139 $\pm$ 0.004} & \bfseries{0.138 $\pm$ 0.002} \\
    & Reddit & $0.146 \pm 0.048$ & $0.270 \pm 0.005$ & $0.198 \pm 0.005$ & $0.072 \pm 0.009$ & \bfseries{0.368 $\pm$ 0.004} & \underline{$0.296 \pm 0.003$} \\
    \midrule
    \multirow{4}{*}{\thead{Top-250 (node proba- \\bility  vs Centrality)}} & DBLP & $-0.094 \pm 0.060$ & $0.151 \pm 0.030$ & $0.053 \pm 0$ & $0.189 \pm 0.038$ & \bfseries{0.323 $\pm$ 0.009} & \underline{$0.307 \pm 0.004$} \\
    & Wikipedia & $0.133 \pm 0.016$ & $0.094 \pm 0.010$ & 0.080 $\pm$ 0.002 & $0.148 \pm 0.019$ & \bfseries{0.289 $\pm$ 0.002} & \underline{0.274 $\pm$ 0.010}\\
     & Reddit & $0.241 \pm 0.014$ & \underline{0.445 $\pm$ 0.016} & $0.200 \pm 0.005$ & $0.124 \pm 0.044$ & \bfseries{0.492 $\pm$ 0.019} & \underline{0.448 $\pm$ 0.011} \\
    \hline
    NMI & DBLP  & $0.401 \pm 0.007$ & \underline{$0.4143 \pm 0.004$} & \underline{$0.4137 \pm 0.0$} & $0.103 \pm 0.007$ & \bfseries{0.429 $\pm$ 0.015} & $0.368 \pm 0.002$ \\
    \bottomrule
  \end{tabular}
 \end{adjustbox}
  \label{community_detection}
\end{table*}

\subsection{Evaluation Metrics}

\textbf{Dynamic link prediction.} 
An important application of dynamic graph embedding is capturing the pattern of temporal edge formation in the training set to predict edges in future time steps. For all baseline methods we use a metric of similarity between node embeddings (Euclidean distance or dot product) as a predictor of connectivity, following each method's implementation. For static methods, we aggregate all observed edges in the training set in a single graph to produce node embeddings. For dynamic baselines, the node representations at the last training step are used. For GRADE, we train our model, and infer the posterior distributions of the node and community embeddings at the test time steps. We do this by using the trained GRUs to produce the node and community embeddings at the future time steps.\\

We evaluate dynamic link prediction performance using the metric \textit{mean average rank} (MAR). To calculate mean average rank we first produce a ranking of candidate neighbours, spanning the complete vertex set, for each source node $v^{test}$ in the test set edge list. Then, we identify the rank of the ground truth neighbour and average over all test edges. The ranking is produced via a similarity measure on the node embeddings for all baseline methods. For GRADE and vGraph, we can use their generative models to produce a distribution over the neighbours for each source node $v^{test}$ in the test set edge list. The procedure for GRADE is as follows: first, we use the trained GRUs to produce the node and community embeddings at test time step $t^{test}$. Then, we use the proposed generative process to produce the parameters of the distributions over the nodes and communities, that is $\theta^{t^{test}}$ and $\pi^{t^{test}}$. Finally, we can produce a distribution over the neighbours by summing over all possible community assignments as follows:
\begin{equation}
p(c \vert v^{test}) = \sum_{z}p_{\theta^{t^{test}}}(c \vert z) p_{\pi^{t^{test}}}(z\vert v^{test}) 
\label{eq: Eq. neighbour_generation}
\end{equation}
We use the node probabilities to produce a ranking of candidate neighbours for source node $v^{test}$. The procedure for vGraph is very similar, the difference being that vGraph cannot generate node and community embeddings at future time steps. This allows us to \textit{directly assess} the benefit of our temporal evolution model in GRADE.\\

\textbf{Dynamic community detection} is another relevant use case for our method. More specifically, we leverage historical information by training a model on the train time steps, and inferring non-overlapping communities given the edges in the test set. For GRADE, this is achieved by inferring the hidden community assignment variable $z$. First, we use the trained GRUs to produce the node and community embeddings at the future time steps, and then we produce community assignments for nodes in the test set using Eq. \ref{eq: Eq. comm_membership_eq}. We evaluate performance on this task using \textit{Normalized Mutual Information} (NMI) \cite{tian2014learning} and \textit{Modularity}. 
Publicly available dynamic network datasets with labelled evolving communities are difficult to obtain. We use the DBLP dataset which we have labelled as described in subsection \textit{A. Data sets}. \\

Further, a novel application of GRADE is \textbf{predicting community-scale dynamics}. We demonstrate this capability by inferring the community representations (i.e., the posterior distribution over the nodes for each community) for the test time steps, and producing rankings of the most probable nodes. A vertex predicted to have high probability for a given community should also be integral to its structure. We evaluate performance on this task by calculating Spearman's rank correlation coefficient between the predicted node probabilities of the top-250 vertices in community $k$ and the same nodes' centrality as measured by the number of links to vertices assigned to the same community $k$ on the test set.

\subsection{Experimental Procedure}

\begin{figure*}[t]
\centering
\includegraphics[width=\linewidth]{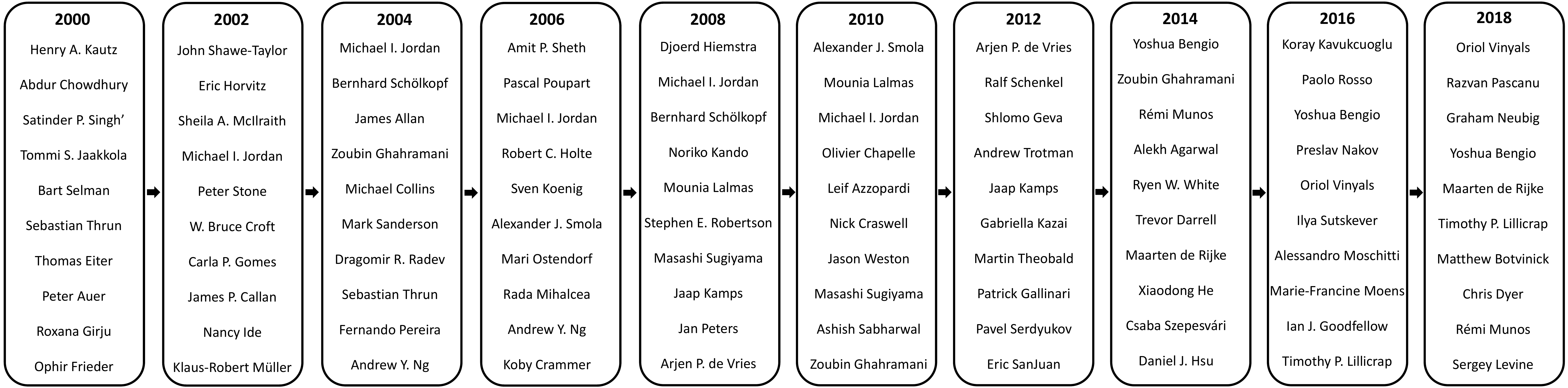}\quad
\caption{Temporal evolution of top-10 authors within a community broadly corresponding to Artificial Intelligence and learnt by GRADE on the DBLP dataset for years 2000-2018 inclusive.}
\label{author_evo_fig}
\end{figure*}

We cross-validate all methods and identify the best set of hyperparameters on the task of dynamic link prediction via grid search. 
The train/validation/test splits are done across time steps as shown in Table~ \ref{dataset_statistics}. We use no node attributes in any of our experiments and set the node embedding dimensionality to 128 for all methods. For all baselines with the exception of vGraph and GRADE, we apply K-means to the learnt vertex representations to identify non-overlapping communities. Further, since the majority of baseline methods (other than vGraph and GRADE) do not produce distributions over the nodes for each community,
we use k-nearest neighbours algorithm to identify the top-250 nodes closest in representation space to each cluster's centroid for the task of predicting community-scale dynamics. For consistency between baselines we determine the number of communities to be detected as part of the cross-validation procedure for GRADE. These were $8$ communities for DBLP, $25$ for Reddit and $50$ for Wikipedia. The implementations provided by the authors are used for all baselines. We train GRADE using the Adam optimizer \cite{kingma2014adam} with an initial learning rate of 0.005 which is decayed by 0.99 every 100 iterations. To save the best models for GRADE and vGraph, we train for 10,000 epochs on the task of dynamic link prediction and select the models with lowest mean average rank on the validation set. The same models were used in the evaluation for all tasks. We use procedures provided by the authors' implementations of DySAT and DynamicTriad to save best-performing models. Owing to the size of dynamic networks we cannot use full-batch training. We resort to training GRADE stochastically by splitting the edges at each time step in equally sized batches comprising $\sim 10,000-75,000$ edges in our experiments. Since our model is transductive we report results on nodes that have been observed in the training set. All results are averaged across 4 runs. Training GRADE on the proposed datasets requires up to $\sim12$ hours on a 12GB NVIDIA TitanX GPU.

\subsection{Results}
\textbf{Results on dynamic link prediction} are shown in table \ref{mars}. GRADE outperforms noticeably on all datasets. More specifically, for DBLP there is a $58\%$ improvement in mean average rank compared to the second-best method, for Reddit - $51\%$ improvement compared to second-best, and for Wikipedia - $70\%$. Often, in real-world graphs the node-degree distribution is highly skewed and nodes cluster around hubs, which are densely connected. These hub nodes give rise to community structure in graphs \cite{barabasi1999emergence}. As GRADE learns to correctly perform community assignment and attribute high probability to hub nodes, we expect GRADE to show favourable performance. This is because the proposed generative mechanism imposes an \textit{inductive bias} on the existence of community structure in the datasets through the hidden community assignment variable $z$. These results demonstrate that introducing an explicit inductive bias about the existence of community structure contributes to the task of dynamic link prediction. Further, projecting the embeddings for future time steps, as opposed to re-using the ones learnt at the last training time step for dynamic prediction as other dynamic baselines do (e.g. DynamicTriad and DySAT), can also be viewed as a performance advantage. \\


\textbf{Results on dynamic community detection} are presented in Table \ref{community_detection}. In this task GRADE also outperforms all baselines on all datasets. Firstly, comparing against vGraph, the significance of injecting temporal evolution in the node and community embeddings can be directly evaluated. The static vGraph cannot capture these dynamics, hence relative improvement of up to $24\%$ for GRADE is observed, depending on the dataset. Secondly, all other baseline methods produce noticeably weaker results even than vGraph. This discrepancy in performance can be attributed to the efficient mechanism for community assignment (see Eq. \ref{eq: Eq. comm_membership_eq}). In contrast, for other baselines (with the exception of vGraph) K-Means clustering on the node embeddings has to be applied to identify communities. Explicitly inferring the hidden community assignment variable is more effective for the task of dynamic community assignment.\\

We also present results on \textbf{predicting community-scale dynamics}. Similarly to other experiments, GRADE also outperforms on this task as can be seen in Table \ref{community_detection}. More specifically, predicting the most structurally significant nodes for future time steps corresponds to correctly identifying the highly probable nodes for the communities. In GRADE, this can be easily accomplished by inferring the community embeddings for future time steps, producing the distributions over the nodes and taking the top 250 nodes. Compared to vGraph, which can also produce node distributions but does not capture time dynamics, GRADE performs better ($5\% - 10\%$ relative improvement). Compared to other baselines, which have to resort to applying the k-nearest neighbours algorithm on the produced node embeddings to find the closest 250 nodes to each cluster centroid, GRADE outperforms substantially. Most notably, dynamic baselines DySAT and DynamicTriad are significantly inferior to GRADE's performance on all datasets. Finally, we also report results on \textbf{NMI on DBLP} in Table \ref{community_detection}. Again, GRADE produces the best results followed by DySAT and node2vec. \\

In Fig. \ref{author_evo_fig}, we visualize the temporal evolution of the top 10 most probable authors from a community strongly associated with Artificial Intelligence, learnt by GRADE on all time steps from DBLP. We observe the top authors in each year work within the same general research area (\textit{coherence}) and the community is broadly in agreement with historical events (\textit{interpretability}). For example, our model assigns high probability to influential researchers like Yoshua Bengio and Ian J. Goodfellow in later years.

\section{Conclusion}
In this paper, we propose GRADE - a method which jointly learns evolving node and community representations in discrete-time dynamic graphs. We achieve this with an edge generative mechanism modelling the interaction between local and global graph structures via node and community multinomial distributions. We parametrize these distributions with the learnt embeddings, and evolve them over time with a Gaussian state-space model. Moreover, we introduce transition matrices to explicitly capture node community dynamics. Finally, we validate the effectiveness of GRADE on real-world datasets on the tasks of dynamic link prediction, dynamic community detection, and the novel task of predicting community-scale dynamics, that is inferring future structurally influential vertices. 

\ifCLASSOPTIONcaptionsoff
  \newpage
\fi
\bibliographystyle{IEEEtran}
%
\vspace{-8pt}
\bibliography{bare_jrnl.bib}
\vspace{-12pt}
\vskip -10pt plus 10pt
\vskip 0pt plus -1fil
\begin{IEEEbiography}[{\includegraphics[width=1in,height=1.25in,clip,keepaspectratio]{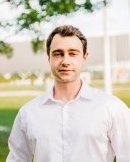}}]{Simeon Spasov}
received his BEng degree in Electrical and Electronic Engineering from the University of Manchester, U.K. in 2015 and his MRes degree in Sensor Technologies and Applications from the University of Cambridge, U.K. in 2016. Currently, he is pursuing a Ph.D. degree in machine learning at the Department of Computer Science and Technology at the University of Cambridge. Mr. Spasov has conducted machine learning research at the Mila-Quebec AI Institute, Montreal, Canada under the supervision of Prof. Jian Tang, as well as in Amazon Alexa. His research interests include machine learning for healthcare and graph representation learning.
\end{IEEEbiography}
\vskip 0pt plus -1fil
\begin{IEEEbiography}[{\includegraphics[width=1in,height=1.25in,clip,keepaspectratio]{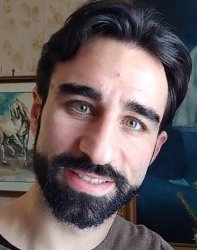}}]{Alessandro Di Stefano} is a Lecturer in Computer Science at Teesside University, within the School of Computing, Engineering and Digital technologies (SCEDT), Department of Computing \& Games. He received his
BSc (2009) and MSc degrees (2012) in Telecommunications
Engineering from the Department of Electrical, Electronic
and Computer Engineering (DIEEI) at the University of
Catania, Italy. He holds a Ph.D. degree in Systems
Engineering from the DIEEI at the University of Catania,
Italy, awarded in 2015. After his Ph.D., he worked as
postdoctoral researcher at DIEEI, University of Catania, and then as a research associate and teaching assistant at the
Department of Engineering, King's College London (KCL),
London, UK, as well as at the Department of Computer Science
and Technology, University of Cambridge, UK. Dr. Di
Stefano's research interests include game theory, network
science and artificial intelligence. He has published many
peer-reviewed papers in high impact journals and leading
international conferences.
\end{IEEEbiography}
\vskip 0pt plus -1fil
\begin{IEEEbiography}[{\includegraphics[width=1in,height=1.25in,clip,keepaspectratio]{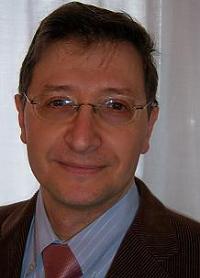}}]{Pietro Li\`o}
received his Ph.D. degree in complex systems
and nonlinear dynamics and an M.A. degree
from the Department of Engineering, School of Informatics, University of Florence, Florence, Italy, in 1995 and 2015, respectively, and a Ph.D. degree in theoretical genetics from the University of Pavia, Pavia, Italy, in 2007. He is currently a Professor in computational biology with the Department of Computer Science and Technology, University of Cambridge, Cambridge, U. K., where he is also a member of the Computer Laboratory, Artificial Intelligence Group. His current research interests include machine learning and data mining, data integration, computational models in health big data, and predictive models in personalized medicine.
\end{IEEEbiography}
\vskip 0pt plus -1fil
\begin{IEEEbiography}[{\includegraphics[width=1in,height=1.25in,clip,keepaspectratio]{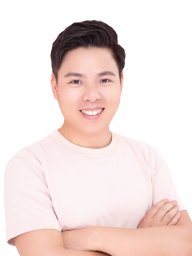}}]{Jian Tang}
is currently an assistant professor at Mila-Quebec AI Institute and HEC Montreal as well as a CIFAR AI Chair. Prior to that, he was a Postdoc at the University of Michigan and Carnegie Mellon University. He also worked at Microsoft Research Asia as an associate researcher between 2014-2016. Prof. Tang obtained his Ph.D. degree in Computer Science from Peking University, Beijing, China in 2014 and his B.S. in Mathematics from Beijing Normal University, Beijing, China in 2009. His research interests include deep learning, graph representation learning, graph neural networks, drug discovery and publishes frequently in top-tier machine learning journals and conferences, such as NeurIPS, ICML, ICLR. Prof. Tang is also the author of the paper "LINE: Large-scale Information Network Embedding" for which he won the Most Cited Paper Award by The World Wide Web Conference in 2015 and holds multiple other awards from companies like Amazon and Tencent.
\end{IEEEbiography}




\end{document}